\documentclass[11pt,a4paper]{article}
\usepackage{crisis-domain}

\usepackage{multicol}
\usepackage{multirow}

\usepackage{xcolor}

\usepackage{amsfonts}
\usepackage{amsmath}

\usepackage{times}
\usepackage{url}
\usepackage{hyperref}
\usepackage{cleveref}

\usepackage{caption}
\usepackage{subcaption}

\usepackage{latexsym}
\usepackage{graphicx}

\usepackage{booktabs}

\usepackage{flushend}

\date{}

\begin{document}
\maketitle

\begin{abstract}

The Incident streams (IS) track is a research challenge aimed at finding important information from social media during crises for emergency response purposes. More specifically, given a stream of crisis-related tweets, the IS challenge asks a participating system to 1) classify what the types of users' concerns or needs are expressed in each tweet, known as the information type (IT) classification task and 2) estimate how critical each tweet is with regard to emergency response, known as the priority level prediction task. In this paper, we describe our multi-task transfer learning approach for this challenge. Our approach leverages state-of-the-art transformer models including both encoder-based models such as BERT and a sequence-to-sequence based T5 for joint transfer learning on the two tasks. Based on this approach, we submitted several runs to the track. The returned evaluation results show that our runs substantially outperform other participating runs in both IT classification and priority level prediction.

\end{abstract}
\section{Introduction}
\label{sec:intro}

Social media platforms such as Twitter have made it possible for users to report on an ongoing event in their vicinity in a timely manner~\cite{fraustino2012social}. This has motivated researchers to explore the potential of social media platforms for finding actionable information from this user-generated content during a crisis event~\cite{caragea2011classifying,imran2015processing,McCreadie2019}. Finding this type of information is especially important for emergency response agencies to enable them to take immediate actions to help those who are posting for help, which is known as situational awareness~\cite{vieweg2012situational,vieweg2010microblogging}. This naturally raises the question: how can the process of finding the actionable information effectively be automated, given the fact that the messages posted during a crisis on social media are usually noisy and numerous?

The Incident streams (IS) track~\cite{McCreadie2019,McCreadie2020} is proposed by the \textbf{T}ext \textbf{RE}trieval \textbf{C}onference (TREC) as a research challenge for this purpose. Since it was introduced in 2018, the IS track has conducted two major tasks regarding crisis short message processing. Given a stream of tweets from crisis events, the foremost task is that it asks a participating system to classify the information types (ITs) for each tweet. The ITs are simply a pre-defined set of classes in relation to something that a user is likely to post during a crisis. The ITs can be something important such as \textit{requesting research and rescue, call for moving people, reporting goods available}, etc., as well as something less important such as \textit{reporting weather or location, expressing sentiment}, etc.\footnote{There are 6 important ITs known as ``actionable'' ITs pre-defined by the IS track and 19 are considered to be ``non-actionable''. For details, see~\cite{McCreadie2019}.} In addition to the ITs classification task, the IS track also asks the participating systems to estimate the priority level for each tweet, indicating how important the tweet is in taking immediate emergency response actions. The IS track pre-defines four priority levels: \textit{critical}, \textit{high}, \textit{medium} and \textit{low}, which are ordered from the highest to lowest priority.

The IS track was run once in 2018 and twice in each subsequent year, so it has accumulated five editions as of 2020. For each edition, an annotated collection of tweets from previous editions is used as the training data for the community, and unseen tweets (non-annotated) are released as the test tweets for official evaluation. The two most recent editions, conducted in 2020, are named 2020A and 2020B respectively. Slightly different from previous editions, the two editions introduce a reduced set of ITs as well as a set of test tweets related to the COVID-19 pandemic, resulting in three tasks described as follows.

\begin{itemize}
    \item \textbf{Task 1}: This task remains the same as the editions before 2020, it uses all 25 ITs for classification and four priority levels for estimation.
    \item \textbf{Task 2}: Different from Task 1, this task only asks the participating systems to classify one or more of 12 IT classes. The 12 ITs include 11 that are closely related to emergency response and the remaining as ``Other-Any''~\footnote{For full details, refer to~\url{http://dcs.gla.ac.uk/~richardm/TREC_IS/2020/participate.html}}.
    \item \textbf{Task 3}: Unlike Task 1 and 2 that relate to general crises such as earthquakes, explosions or hurricanes, this task focuses on the COVID-19 domain. It provides a stream of COVID-related tweets from different locations for IT classification using only a subset of 9 ITs suitable for COVID-19 and priority estimation using the same four priority levels as used in Task 1 and 2.
\end{itemize}

In this paper, we describe our system's approach in the three tasks of the IS track from our participation in both 2020A and 2020B. For different tasks, we submitted different runs but all were based on the multi-task transfer learning approach that we utilised in our system. Given the recent success of transformers~\cite{Wolf2019HuggingFacesTS} in transfer learning for various language tasks such as sentence classification, question answering, etc., we leverage them in the IS challenge. We explored transformer encoder based models such as BERT~\cite{BERT2018} and a sequence-to-sequence model - T5~\cite{raffel2019exploring} for their potential in this challenge. By doing so, we fine-tune them in a multi-task learning fashion (i.e, joint fine-tuning of the IT classification and priority estimation). With this approach, we submitted five runs to the IS track. The evaluation results show that our runs substantially outperform other participating runs in both IT classification and priority level prediction.

\section{Related Work}

To improve emergency response, the community has seen many works on exploring computational techniques for knowledge acquisition from crisis messages on social media. \citet{caragea2011classifying} applied traditional machine learning algorithms including LDA and SVM to find important information such as \textit{people trapped} or \textit{food shortage} from the 2010 Haiti Earthquake. As neural network (NN) approaches have gained popularity in recent years, many deep learning approaches have been applied to this domain. For example,  \citet{nguyen2017robust} applied a convolution neural network (CNN) for classifying informative tweets from general disasters such as the \textit{2015 Nepal Earthquake}, \textit{Typhoon Hagupit}, etc., whereas~\citet{alam2018domain} leveraged a CNN with adversarial training for identifying whether a tweet is relevant to a certain crisis event. 

In recent years, since the attention-based transformer model was introduced~\cite{vaswani2017attention}, several variations have been proposed such as BERT~\cite{BERT2018}, ELECTRA~\cite{Clark2020ELECTRAPT} and T5~\cite{raffel2019exploring}, collectively known as the transformers~\cite{Wolf2019HuggingFacesTS}, achieving state-of-the-art performance in many language tasks with transfer learning. It is common that the transformers are first pre-trained on a large general text corpus and then are fine-tuned on specific downstream language tasks such as text classification. Given the strong transfer capability of transformers, they have been widely studied for crisis messages processing also. \citet{liu2020crisisbert} fine-tuned BERT for crisis identification and detection tasks and \citet{wang2020ucdcs} applied T5 for extracting useful information such as \textit{who tested positive/negative or cannot get test} from COVID-related tweets by treating it as a question-answering task. Our approach in the IS track is similar to this line of work, which applies the transformers with transfer learning for finding actionable information in the tasks as proposed by the IS track. However, our approach is different in the way it fine-tunes the transformers by multi-task learning, aiming to make use of shared model weights between different tasks. 

Since the IS track has been run for several years, the participating systems have proposed various techniques specifically for this track. Such approaches can broadly be summarised in three categories. First, traditional machine learning algorithms have been used with careful pre-processing steps and handcrafted input features. For example, \citet{wangcmu} applied models including Na\"ive Bayes, SVM, Random Forest, and the ensemble of these models. To train these models, they used hand-crafted features such as the length, sentiment polarity of a tweet, number of followers of the user, combining with context-free GloVe and FastText embeddings as well as context-aware BERT embeddings as the input features. The second category uses deep learning approaches that pre-date the widespread adoption of transformers. For instance, \citet{miyazaki2019label} proposed the method using label embedding with a BiLSTM model in this track while \citet{congcong2020cls} applied a BiLSTM network along with pre-trained ELMo embeddings and trainable embeddings as the input features for crisis tweet categorisation. The last category encompasses transformer-based fine-tuning approaches. One example is that \citet{zahera2019fine} fine-tuned BERT for the multi-label ITs classification task using the training tweets after preprocessing.

\section{Method}
\label{sec:method}

 Our approach is based on multi-task transfer learning through fine-tuning both transformer encoder-based models such as BERT and sequence-to-sequence transformers such as T5. The following details the process of the two types of models used, which we name the \textbf{encoders scenario} and \textbf{sequence-to-sequence scenario} respectively. Each type of model was used for both the IT classification task and priority prediction task.

\textbf{Encoders scenario}: This scenario simply adds two linear projection layers on top of transformer encoders such as BERT. Our architecture is agnostic as to the specific transfer encoder used. One projection layer transforms the encoder's pooled output (namely, the \texttt{[CLS]} output vector of BERT) to a vector representing the IT classes. The IT representation is then passed to the \textit{sigmoid} function that calculates the probability distribution for every IT class. The other projection layer is used to transform the encoder's output to a vector representing the four priority levels. Similarly, it is then passed to the \textit{sigmoid} function, which calculates a score indicating the priority levels as follows.

\begin{equation}
\label{eq:pri-schema}
   (0.75,1] \longrightarrow \mathbf{Critical} 
\end{equation}
$$(0.5,0.75] \longrightarrow \mathbf{High}$$
$$(0.25,0.5] \longrightarrow \mathbf{Medium}$$
$$[0.0,0.25] \longrightarrow \mathbf{Low}$$

In order to achieve the joint learning of both tasks, the encoder model is fine-tuned with the loss function linearly combining the binary cross entropy loss between the IT probability distribution and ground truths (a multi-label classification problem) as well as the mean squared error between the importance scores and priority ground truths (a regression problem).

\textbf{Sequence-to-sequence scenario} (seq2seq): This scenario is mostly motivated by the work that applies T5 for COVID-related event extraction by treating it as a multi-choice question answering task~\cite{congcong2020cls}. We adapt it to the IS track for multi-task transfer learning using seq2seq transformers such as T5. Basically, the seq2seq model takes a sequence of text as the input, known as the source sequence, and outputs the target sequence conditional on the source sequence. Under this mechanism, the template used to construct the source and target sequences in both tasks of the IS track is presented as follows.  

\begin{small}
\vspace{0.2cm}
\textbf{Source}: \colorbox{lightgray}{context: T question: IQ/PQ choices: IC/PC}

\textbf{Target}: \colorbox{lightgray}{I/P}
\end{small}

\begin{itemize}
    \item \textbf{T} refers to the raw tweet text without any reprocessing except for being lower-cased.
    \item \textbf{IQ/PQ} refers to the IT classification and priority estimation task-specific ad-hoc question texts, which are ``\textbf{what type of information does the tweet convey relating to a crisis?}'' and ``\textbf{what level of urgency is likely expressed in this tweet relating to a crisis?}'' respectively.
    \item \textbf{IC/PC} implies the flatted texts concatenating all IT and priority levels respectively. For example, IC is something like ``\textbf{call for donations, call to move people, ...}'' which varies in different IT classification tasks. The PC is simply ``\textbf{critical, high, medium, low}''.
    \item \textbf{I/P} indicates the generated predictions for ITs and priority level, which are direct textual predictions from \textbf{IC/PC} respectively.
\end{itemize}

\begin{table*}[!t]
    \centering
    \begin{tabular}{c|c|c|c|c}
    \toprule
         runtag & scenario & task target & submission type & training data \\
         \midrule
         run1 & Encoders & Task 1 \& 2 & one-hot & prior to 2020B excluding COVID\\
         run2 & Encoders & Task 1 \& 2 & probability& prior to 2020B excluding COVID\\
         run3 & seq2seq & Task 1 \& 2 & one-hot & prior to 2020B excluding COVID\\
         run4 & seq2seq & Task 3 & one-hot& prior to 2020B including COVID\\
        run5 & seq2seq & Task 1 \& 2 & one-hot& prior to 2020B including COVID\\
          \bottomrule  
    \end{tabular}
    \caption{The summary of our submitted runs for TREC-IS 2020-B. Run1, 2, 3, 5 submitted to task 1 are also submitted to task 2 for evaluation.}
    \label{tab:submitted-runs}
\end{table*}

Using this template, each tweet in the training set is converted to an IT-specific source-target pair and a priority-specific source-target pair. In order to achieve the joint learning of both tasks, the sequence-to-sequence model is fine-tuned on batches of training sequences that contain both the IT pairs and priority pairs.

\section{Experiments}

This section describes the details of our system's runs submitted to the latest 2020B edition of the IS track. 
Since our system was developed based on our previous experience in this track, the method we described in Section~\ref{sec:method} also covers our approach to the 2020A edition (actually the \textbf{encoders scenario}). Our baseline run (run1) for 2020B, which is an ensemble run under the \textbf{encoders scenario} from 2020A that we consider as a strong baseline. In 2020B, we submitted a total of five runs to Task 1, 2 and 3 as mentioned in Section~\ref{sec:intro} and they are summarised in Table~\ref{tab:submitted-runs} and described as follows.

\begin{itemize}
    \item \textbf{run1}: This is a baseline with techniques initially developed in 2019A. In 2020A, we proposed the \textbf{encoders scenario}, achieving strong performance as compared to other participating techniques. To further make it a strong baseline, we used a simple ensemble approach combining the predictions made by the fine-tuned individual models\footnote{The individual models that were used in this run included fine-tuned \texttt{bert-base-uncased}, \texttt{electra-base-discriminator}, \texttt{albert-\-base\-v2} and \texttt{distilbert-base-uncased}, which are all available in the transformers library~\cite{Wolf2019HuggingFacesTS}.} under the encoders scenario. The ensemble run simply predicts the final IT predictions for each tweet to be the union of individual IT predictions and the final priority level to the highest of the individual priority predictions. Per the guideline of 2020B, both the IT and priority levels are expected to be numeric instead of being categorical as required prior to 2020B. Hence, we transform the final IT predictions to one-hot encodings and map the priority level prediction to its importance score by: \textit{Critical: 1.0, High: 0.75, Medium: 0.5, Low: 0.25}.
    \item \textbf{run2}: Similar to run1, the difference is that for run2, the final ITs predictions are the highest probability values among the predictions by individual models. The final priority predictions are simply the highest of the individual models' outputs without applying the conversion as defined in Equation~\ref{eq:pri-schema}. 
    \item \textbf{run3}: For this run, the \textbf{seq2seq scenario} is conducted for multi-task transfer learning. We follow the T5 base architecture initialised with \texttt{t5-base} weights and fine-tune it on the training tweets prior to 2020B (excluding the COVID-related tweets from the 2020A edition). Since the seq2seq model outputs the generated texts as the predictions for both priority and ITs, we convert the IT predictions to one-hot encodings and priority level to the importance score before they are submitted.
    \item \textbf{run4}: With a similar setup to run3, run4 is submitted for Task 3 and thus it includes the training tweets prior to 2020B including the COVID-related tweets from 2020A.
    \item \textbf{run5}: With a similar setup to run3, run5 is submitted for Task 1 \& 2 and it uses all previous training tweets including the COVID tweets for fine-tuning the T5 model.
\end{itemize}

\begin{table*}[!h]
\small
\centering
\begin{tabular}{llp{2cm}p{1.5cm}p{1.5cm}p{1.9cm}p{1.5cm}}

\toprule
Run             & nDCG@100        & Info-Type    F1 {[}Actionable{]} & Info-Type F1 {[}All{]} & Info-Type Accuracy & Priority F1 {[}Actionable{]} & Priority F1 {[}All{]} \\
\midrule

BJUT-run        & 0.4346          & 0.0266          & 0.0581          & 0.8321          & 0.1744          & 0.0905          \\
njit.s1.aug     & 0.4480          & 0.2634          & 0.3103          & \textbf{0.8655} & 0.2029          & 0.1518          \\
njit.s2.cmmd.t1 & 0.4475          & 0.1879          & 0.2223          & 0.8475          & 0.2029          & 0.1518          \\
njit.s3.img.t1  & 0.4222          & 0.1879          & 0.2223          & 0.8475          & 0.1959          & 0.1417          \\
njit.s4.cml.t1  & 0.4164          & 0.1712          & 0.1465          & 0.8445          & 0.1054          & 0.1064          \\

ufmg-sars-test  & 0.3634          & 0.0001          & 0.0493          & 0.8337          & 0.1285          & 0.1378  \\
\midrule
ucd-run1 (ours)        & 0.5033          & \textbf{0.3215} & \textbf{0.3810} & 0.8520          & 0.2582          & 0.2009          \\
ucd-run2 (ours)        & 0.5022          & 0.3078          & 0.3692          & 0.8316          & 0.2582          & 0.2016          \\
ucd-run3 (ours)        & 0.5038          & 0.3001          & 0.3448          & 0.8653          & \textbf{0.2803} & 0.3046          \\
ucd-run5 (ours)        & \textbf{0.5252} & 0.3036          & 0.3444          & 0.8601          & 0.2801          & \textbf{0.3126} \\

\bottomrule
\end{tabular}
\caption{Evaluation results of participating runs at TREC-IS 2020-B Task 1. Highest in columns are bold.}
\label{tab:res-task1}
\end{table*}

\begin{table*}[]

\centering
\begin{tabular}{lllll}
\toprule
Run               & nDCG@100        & Info-Type F1 {[}All{]} & Info-Type Accuracy & Priority F1 {[}All{]} \\
\midrule\midrule
                  & \multicolumn{4}{c}{Task-1 Systems} 
                  \\
BJUT-run        & 0.4350          & 0.0472          & 0.7977          & 0.1337          \\
njit.s1.aug     & 0.4487          & 0.3480          & 0.8846          & 0.1838          \\
njit.s2.cmmd.t1 & 0.4467          & 0.2494          & 0.8612          & 0.1838          \\
njit.s3.img.t1  & 0.4215          & 0.2494          & 0.8612          & 0.1708          \\
njit.s4.cml.t1  & 0.4176          & 0.1278          & 0.8360          & 0.1162          \\

ufmg-sars-test  & 0.3630          & 0.0127          & 0.8419          & 0.1480     \\
\midrule
ucd-run1 (ours)        & 0.5020          & \textbf{0.4036} & 0.8913          & 0.2320          \\
ucd-run2 (ours)        & 0.5027          & 0.3961          & 0.8364          & 0.2322          \\
ucd-run3 (ours)        & 0.5032          & 0.3689          & \textbf{0.8932} & 0.2867          \\
ucd-run5 (ours)        & \textbf{0.5240} & 0.3674          & 0.8845          & \textbf{0.3003} \\
\midrule\midrule
                  & \multicolumn{4}{c}{Task-2 Systems}                                          \\
njit.s1.aug.t2    & 0.4478 & 0.2548 & 0.8656 & 0.1838 \\
njit.s2.cmmd.t2   & 0.4478 & 0.2548 & 0.8656 & 0.1838 \\
njit.s3.img.t2    & 0.4213 & 0.2548 & 0.8656 & 0.1708 \\
njit.s4.cml.t2    & 0.4189 & 0.1713 & 0.8327 & 0.1162 \\
ufmg-sars-test-t2 & 0.3637 & 0.0127 & 0.8419 & 0.1480 \\
\bottomrule

\end{tabular}

\caption{Evaluation results of participating runs at TREC-IS 2020-B Task 2. The Task-1 systems refer to the runs from Task 1 re-evaluated under Task 2 while Task-2 systems are the submitted runs specific to Task 2.}
\label{tab:res-task2}
\end{table*}

\subsection{Training Details}

As described, our runs mainly focus on fine-tuning several transformer encoder models and a \textbf{t5-base} sequence-to-sequence model in a multi-task learning way. For the fine-tuning of \textbf{t5-base}, we follow the same hyper-parameter configuration as used in~\citet{wang2020ucdcs}. For fine-tuning each of the transformer encoder models, we use the same set of the hyper-parameters that are configured with reference to a similar work in this domain~\cite{liu2020crisisbert}. For training, we sample around 10\% of the training data as the validation set first. Then, we fine-tune each model with a batch size of 32, learning rate of 5e-5, linear warm-up ratio of 0.1 with Adam optimizer~\cite{kingma2014adam}. For the input length, we set the maximum input length to be 256 since we found few examples has length beyond this number. All training examples in our experiments are not pre-processed but used in raw texts.

\subsection{Results}

\begin{figure*}[!h]
     \centering
     \begin{subfigure}[b]{0.45\textwidth}
         \centering
         \includegraphics[width=\textwidth]{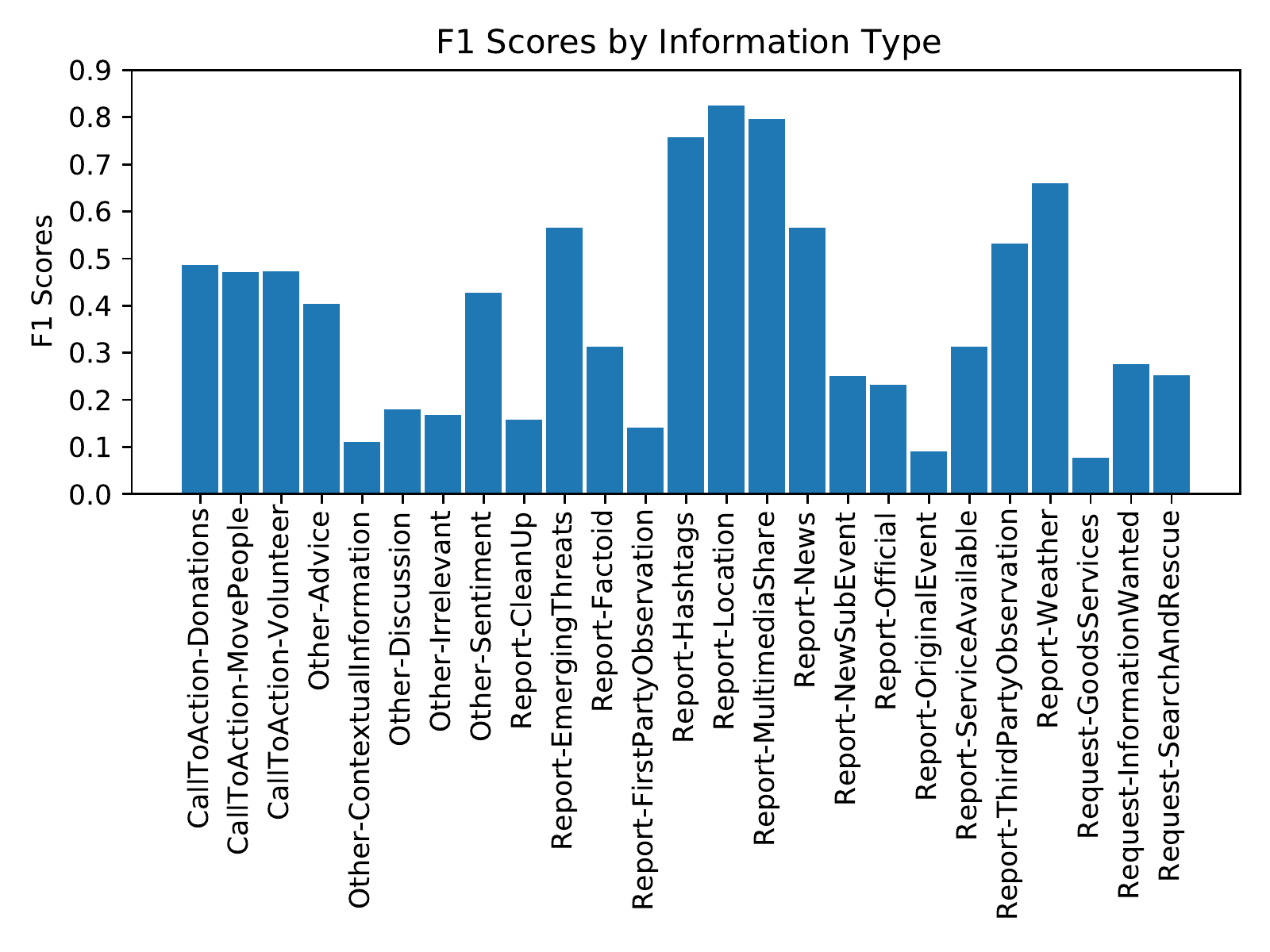}
         \caption{F1 Scores by Information Type}
         \label{fig:scores-per-it}
     \end{subfigure}
     \begin{subfigure}[b]{0.45\textwidth}
         \centering
         \includegraphics[width=\textwidth]{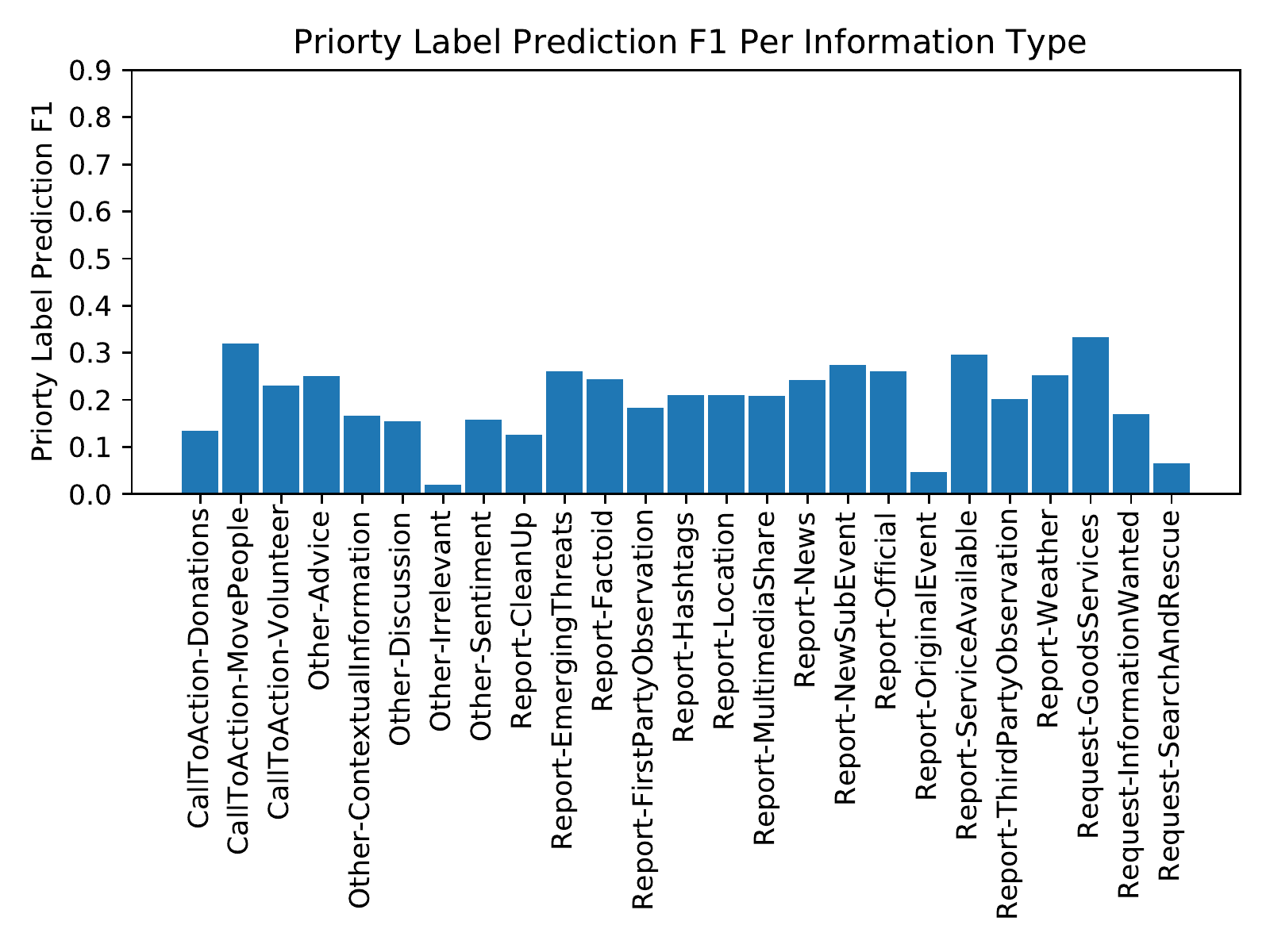}
         \caption{Priority level prediction F1 per information type}
         \label{fig:pri-per-it}
     \end{subfigure}
        \caption{Performance visualisation by information types of \textbf{ucd-run1} in Task 1.}
        \label{fig:perf-vis-run1}
\end{figure*}
\begin{table*}[]
\small
\centering
\begin{tabular}{llp{2cm}p{1.5cm}p{1.5cm}p{1.7cm}p{1.5cm}}
\toprule
Run             & nDCG@100        & Info-Type    F1 {[}Actionable{]} & Info-Type F1 {[}All{]} & Info-Type Accuracy & Priority F1 {[}Actionable{]} & Priority F1 {[}All{]} \\
\midrule
njit.s1.aug.t3  & 0.4322          & \textbf{0.1629} & 0.1450          & 0.8593          & 0.2551          & 0.1499          \\
njit.s2.cmmd.t3 & 0.4329          & 0.1590          & 0.1184          & 0.8586          & 0.2551          & 0.1499          \\
njit.s3.img.t3  & 0.3986          & 0.1590          & 0.1184          & 0.8586          & 0.2544          & 0.1562          \\
njit.s4.cml.t3  & 0.4249          & 0.0210          & 0.0650          & \textbf{0.8626} & 0.1375          & 0.1502          \\
\midrule
ucd-run4        & \textbf{0.4497} & 0.1425          & \textbf{0.1817} & 0.8541          & \textbf{0.3443} & \textbf{0.2867} \\
\bottomrule
\end{tabular}
\caption{Evaluation results of participating runs at TREC-IS 2020-B Task 3.}
\label{tab:res-task3}
\end{table*}

Having submitted the five runs as described in Table~\ref{tab:submitted-runs} to the track, they were officially evaluated and the results are reported in Tables~\ref{tab:res-task1}, \ref{tab:res-task2} and~\ref{tab:res-task3}. The tables show the performance of participating runs in Task 1, 2 and 3 respectively. The columns are the official metrics used to evaluate different aspects of a run's performance, which are described briefly as follows.

\begin{itemize}
    \item \textbf{Information type classification}: There are two types of information type (IT) F1. The ``Actionable IT'' F1 reflects a run's performance in classifying actionable ITs~\footnote{They are Request-GoodsService, Request-SearchAndRescue, Report-NewSubEvent, Report-ServiceAvailable, CallToAction-MovePeople, and Report-EmergingThreats.}. The ``All IT'' F1 measures a run's performance across all information types (25 in Task 1, 12 in Task 2 and 9 in Task 3). The IT accuracy is the overall accuracy in  IT classification.
    \item \textbf{Prioritisation}: Similarly, the Actionable priority F1 measures a run's performance in priority level prediction for only the tweets that are labeled as actionable ITs while the All F1 measures the performance for all test tweets. Moreover, the nDCG@100 is used to measure a run's average performance in ranking top 100 test tweets per event by priority.
\end{itemize}

As seen from Table~\ref{tab:res-task1}, in Task 1, our runs substantially outperform other participating runs in both IT classification and prioritisation~\footnote{The exception if accuracy, where only a small difference is observed accuracy across the participating runs: our results are substantially higher than other participating runs in the remaining metrics.}. In particular, our runs are effective in classifying actionable ITs. For example, our run1 and run3 achieve the top actionable IT F1 score of 0.3215 and the best actionable priority F1 of 0.2803 respectively. This is further evidenced by the runs' performance in Task 2, as in Table~\ref{tab:res-task2}. All the runs overall perform well in IT classification and prioritisation in Task 2 (the condensed more emergency response related 12 ITs). 

In Task 1 and 2, run1 and run2 perform similarly across the metrics since both are based on the encoder scenario and only differ in the final submission type. It is interesting that run5 performs similarly to run3 across the metrics except for being better in nDCG@100: 0.5252 versus 0.5038. The two runs are both based on the seq2seq scenario and only difference is in their training data. This indicates that adding the COVID data (similar domain) to the general crisis data for training can be helpful in the priority-centric ranking performance. To compare between the four runs, it is found that no one run dominates the other runs across all the metrics. This indicates that the multi-task transfer learning approach using either the transformer encoder or the seq2seq as the base model is likely to bring similar performance. 

To further examine our runs' performance at every IT level, we report the IT F1s and priority F1s per IT of the run1 in Task 1, as presented in Figure~\ref{fig:perf-vis-run1}. Figure~\ref{fig:scores-per-it} shows that the run performs well in categorising some actionable ITs, such as ``CallToAction-MovePeople'' and ``Report-EmergingThreats'' while not the best in actionable ITs such as ``Request-GoodsService'', as compared to the non-actionable ITs. However, taking a look at the priority F1s per IT in Figure~\ref{fig:pri-per-it}, we found that the run performs relatively better in priority level prediction for actionable ITs than non-actionable ITs, where ``CallToAction-MovePeople'', ``Request-GoodsService'' and ``Report-ServiceAvailable'' are the top 3 the runs achieve in priority F1. 

Apart from the four runs to Task 1 and Task 2, we submitted run4 to Task 3 and the results are reported in Table~\ref{tab:res-task3}. We see that the run is competitive with other participating runs, particularly in prioritisation. Unlike our other four runs in Task 1 and 2, this run achieves 0.1425 in actionable IT F1, next to the best 0.1629. Since Task 3 is COVID-related and newly introduced, we expect our run to be improved in future iterations of this track as more data accumulates. 

\section{Conclusion}

This paper introduces University College Dublin's (UCD) participation in the 2020 TREC-IS track. The IS track was run twice in 2020: namely 2020A and 2020B. Based on our experience from previous editions, we describe our multi-task transfer learning approach using pre-trained encoder-based and sequence-to-sequence transformers. With these approaches, we submitted five runs to the track's 2020-B edition - four for Task 1 and Task 2, and one for Task 3. The results show that our runs to Task 1 and Task 2 substantially outperform other participating runs in both information type classification and priority level prediction. In addition, our runs are effective in finding some actionable information types in Task 1 and Task 2 and the run to Task 3 performs competitively with other participating runs. Regarding future work, we expect to explore the incorporation of knowledge graphs to enhance the model's identification of the crisis-related tweets.

\bibliography{crisis-domain}
\bibliographystyle{acl_natbib} 
\appendix
\end{document}